\documentclass[10pt,twocolumn,letterpaper]{article}
\usepackage{graphics} % for pdf, bitmapped graphics files
\usepackage{epsfig} % for postscript graphics files
\usepackage{mathptmx} % assumes new font selection scheme installed
\usepackage{times} % assumes new font selection scheme installed
\usepackage{amsmath} % assumes amsmath package installed
\usepackage{amssymb}  % assumes amsmath package installed
\usepackage{graphicx}
\usepackage[bookmarks=false]{hyperref}
\usepackage{cite}
\usepackage{footnote}
\usepackage{amsmath,amssymb,amsfonts}
\usepackage{algorithmic}
\usepackage{textcomp}
\usepackage{pdflscape}
\usepackage{tabularx}
\usepackage{subcaption}
\usepackage{amsmath}
\usepackage{mathtools}
\usepackage{tabu}
\usepackage{booktabs}% for better rules in the table
\usepackage{lipsum}
\usepackage{rotating}
\usepackage{btas}

\usepackage{times}
\usepackage{epsfig}
\usepackage{graphicx}
\usepackage{amsmath}
\usepackage{amssymb}
\usepackage{array}
\usepackage{multirow}
\usepackage{fancyhdr}

\newcommand\MyBox[2]{
  \fbox{\lower0.75cm
    \vbox to 1.7cm{\vfil
      \hbox to 1.7cm{\hfil\parbox{1.4cm}{#1\\#2}\hfil}
      \vfil}%
  }%
}

\btasfinalcopy % *** Uncomment this line for the final submission

\ifbtasfinal\fi

\begin{document}

%%%%%%%%% TITLE
\title{On Detecting GANs and Retouching based Synthetic Alterations}

\author{Anubhav Jain, Richa Singh, Mayank Vatsa\\
IIIT Delhi, India\\
% New Delhi, India\\
{\tt\small \{anubhav15129, rsingh, mayank\}@iiitd.ac.in}
% \and
% Richa Singh\\
% IIIT Delhi\\
% New Delhi, India\\
% {\tt\small rsingh@iiitd.ac.in}
% \and
% Mayank Vatsa\\
% IIIT Delhi\\
% New Delhi, India\\
% {\tt\small mayank@iiitd.ac.in}
}

\maketitle

\thispagestyle{fancy}

%%%%%%%%%%%%%%%%%%%%%%%%%%%%%%%%%%%%%%%%%%%%%%%%%%%%%%%%%%%%%%%%%%%%%%%%%%%%%%%%
\begin{abstract}

Digitally retouching images has become a popular trend, with people posting altered images on social media and even magazines posting flawless facial images of celebrities. Further, with advancements in Generative Adversarial Networks (GANs), now changing attributes and retouching have become very easy. Such synthetic alterations have adverse effect on face recognition algorithms. While researchers have proposed to detect image tampering, detecting GANs generated images has still not been explored. This paper proposes a supervised deep learning algorithm using Convolutional Neural Networks (CNNs) to detect synthetically altered images. The algorithm yields an accuracy of 99.65\% on detecting retouching on the ND-IIITD dataset. It outperforms the previous state of the art which reported an accuracy of 87\% on the database. For distinguishing between real images and images generated using GANs, the proposed algorithm yields an accuracy of 99.83\%. 

\end{abstract}

%%%%%%%%%%%%%%%%%%%%%%%%%%%%%%%%%%%%%%%%%%%%%%%%%%%%%%%%%%%%%%%%%%%
\section{Introduction}
Digital images have become an essential part of our daily lives. With the availability of sophisticated image processing tools and techniques, the Internet is filled with fake images. While some of these images are harmless, others have been used for creating forged legal documents, presenting doctored evidence in court, and manipulating historic incidences. Today, social media is also flooded with re-touched images which makes someone's skin look flawless. Social media websites have also started promoting retouching, by introducing image filters to enhance your appearance. These filters let the user remove wrinkles, pimples, change basic facial structures, and add texture, along with altering facial color i.e. forging skin tones to have fairer skin or adding unnatural tanning effect. Similarly, as shown in Figure 1, beauty or celebrity magazines which are giving people unrealistic expectations with altered appearances.  

One of the most challenging aspects of image forgery is that, when done carefully, it can be visually imperceptible. Russello \cite{russello2009impact} showed that such altered appearances lowers self-esteem of people trying to adhere to the societal norms on attractiveness. It also leads to body dissatisfaction due to unrealistic body images being portrayed. In 2013, Israel announced its plans to enact the “Photoshop Law”. This law makes it mandatory for advertisers and magazines to label photo-shopped images \cite{photoshop}. It shows the necessity of algorithms to detect tampering and also the extent to which this issue is prevalent.

\begin{figure}[tbp]

\centerline{\includegraphics[width=0.8\columnwidth]{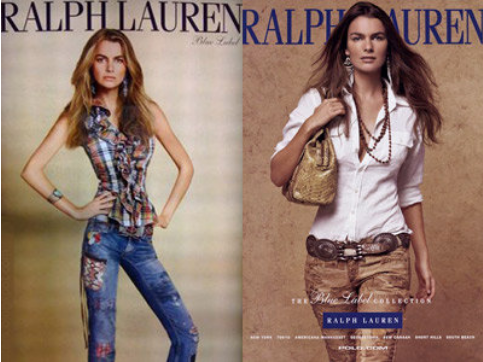}}
\caption{The image on the left is the altered body image of model Countess Filippa Hamilton and the image on right is an unaltered image. Image taken from \cite{ralph}}
\label{fig1}
\end{figure}

Apart from the health and moral effects of image retouching, synthetic alterations also affect biometric system used for identification of individuals. There is a plausibility that the doctored image might be unrecognizable or incomparable with its original version. This could hinder the identification process or automatic matching with original faces. Recent studies have shown that face recognition models suffer in the presence of retouching or make-up \cite{impact_biometric}. While facial images are being used in identification cards, there is a need for an automatic system which can detect retouched images.

% \begin{figure}[htbp]

% \centerline{\includegraphics[width=\columnwidth]{images2.jpg}}
% \caption{The first row consists of original images and the second row consists of their respective retouched versions}
% \label{fig}
% \end{figure}

\begin{figure}[t!]

\centerline{\includegraphics[width=0.8\columnwidth]{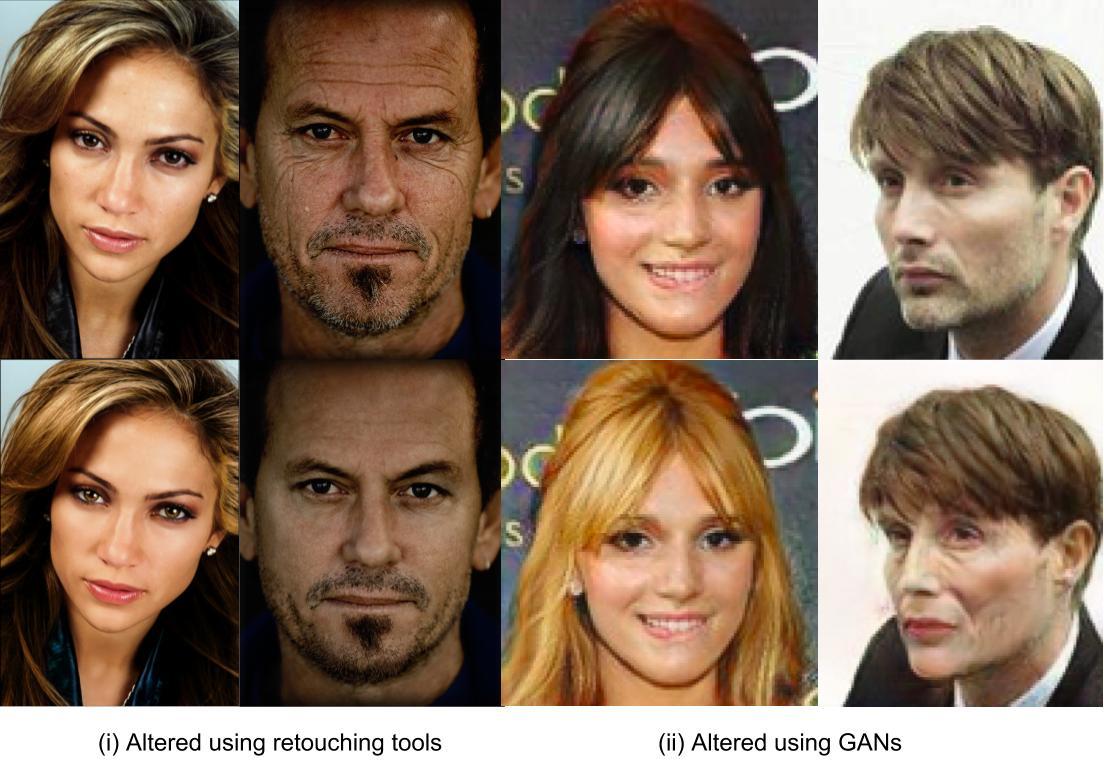}}
\caption{The first row consists of original images and the second row consists of their altered approaches \cite{image2, jennifer}. The first two samples show retouching and the remaining two samples are generated using StarGANs.}
\label{fig:stargan}
\end{figure}

%\subsection{Synthetic alterations using GANs}
More recently, with the emergence of Generative Adversarial Networks \cite{choi2017stargan, pix2pix, cyclegans}, researchers have been exploring image generation as well. With these algorithms becoming more sophisticated, generated images are now looking exceptionally realistic. Various GANs such as CycleGANs \cite{cyclegans} are used for learning image to image translations. Pix2pix GANs \cite{pix2pix} network is able to generate images using label maps. It also uses an image to image translation approach and is able to color images and even generate images from edge maps. While the use of generated images has not been reported for manipulation, it possesses similar powers as tampered images, if not more. This makes it important to also have a mechanism to detect such “fake” images. 

\begin{figure*}[t]
  \centering
\includegraphics[width=18cm, height=6.7cm]{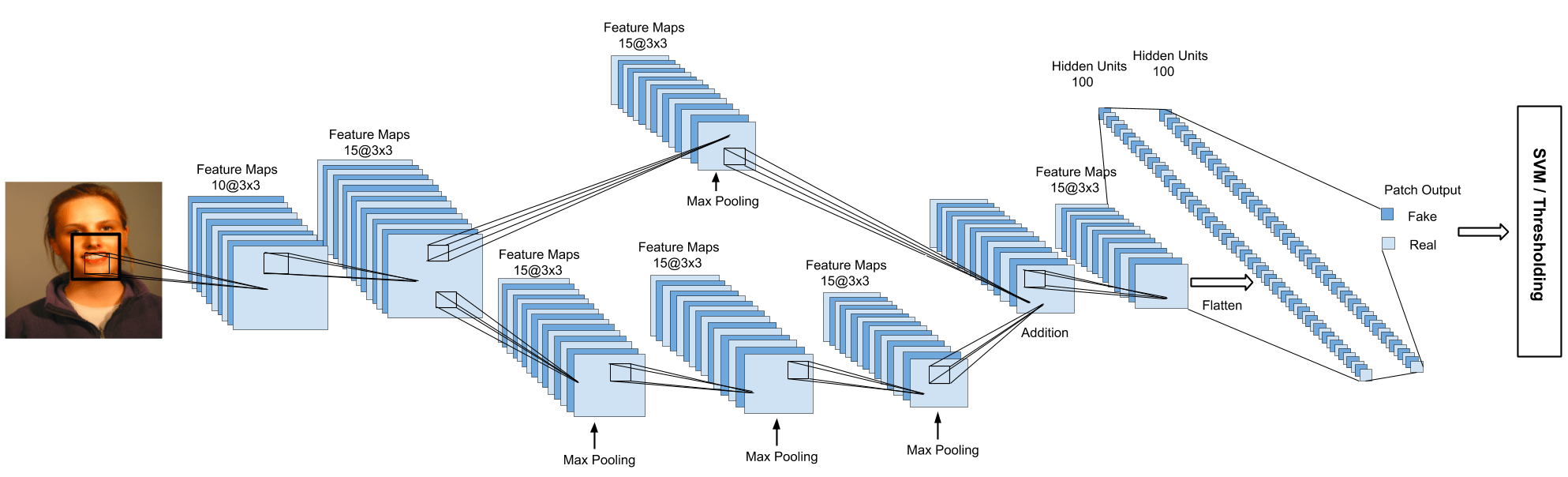}
\label{fig:pipeline}
  \caption{Illustrating the steps involved in the Convolutional Neural Network Architecture for alteration detection.}
%  \hspace*{\fill}
\end{figure*}

\subsection{Related Work}
Retouching, makeup detection, face spoofing and morphing are widely studied areas, that can be considered similar to retouching detection. Recent work by Bharati et al. \cite{7464282} makes use of supervised deep Boltzmann machine algorithm for detecting retouching on the ND-IIITD database. It also introduces the ND-IIITD dataset which consists of 2600 original and 2275 retouched images. It uses different facial parts to learn features for classification. In 2017, Bharati et al. \cite{DBLP:journals/corr/abs-1709-07598} proposed an algorithm which uses semi-supervised autoencoders. The paper has reported results on the Multi-Demographic Retouched Faces (MDRF) dataset. Earlier research by Kee and Farid \cite{Kee19907} learned a support vector regression (SVR) between the retouched and original images. They used both geometric and photometric features for training the SVR on various celebrity images. 

Research in the broad area of facial image forensics, include the paper by Kose et al. \cite{kose} which uses SVM and alligator classifier on a feature vector consisting of shape and texture characteristics. The have reported accuracies for makeup detection on the YMC and VMU datasets. %Boulkenafet \cite{boulkenafet2016face} proposed an algorithm for detecting color textures for face spoofing. The algorithm uses feature description of the luminance and chrominance channels. It reports accuracies on the CASIA anti-spoofing database, replay attack database and MSU mobile face spoofing database. 
Singh et al. \cite{singh2013face} presents an algorithm which detects tampered face images. The algorithm makes use of a gradient based approach for classification. 

\subsection{Contributions}

While existing research has primarily focuses on one of the challenges, this paper proposes a convolutional neural network based algorithm to detect retouching and image generation using GANs. The results are demonstrated using images generated from StarGAN \cite{choi2017stargan} and facial retouching on the ND-IIITD dataset \cite{7464282}. 

\section{Proposed Detection Algorithm}

Different kinds of tampering/retouching algorithms introduce different kinds of irregularities in the face image. As the alterations visually blend inside the image there is a need to focus on local regions, boundary regions and texture. Convolutional neural networks such as ResNet \cite{szegedy2017inception} have demonstrated effective results for different image classification challenges by encoding local and global features. Therefore, we have proposed CNN based architecture for detecting alterations.

%Researchers have shown interest in convolutional neural networks (CNN) for a very long time based on its success with various classification tasks. Along with this, ResNet \cite{szegedy2017inception} showed promising results on classification of images in the ImageNet dataset \cite{Imagenet}. This is the motivation behind using a residual connection based convolutional neural network. As the alterations visually blend inside the image there is a need to focus on local regions, boundary regions and texture. For this purpose, a patch based approach is used which focuses on such features.  

\subsection{Convolutional Neural Network}

As shown in Figure 3, the proposed approach is built on the CNN architecture, where the first step consists of extracting non-overlapping patches of size (64,64,3) or (128,128,3) (only in the case of detecting retouching) from the image. The extracted patches are used as an input for the convolutional neural network which detects various features such as edges, texture and objects. The architecture consists of 6 hidden convolutional layers and 2 fully connected layers.  The convolutional layers use kernel size of (3,3,D) where D is the depth of the filter.

%\subsubsection{Residual Connection}

Inspired from the wider architecture of ResNet \cite{zagoruyko2016wide} and the residual connections, the proposed algorithm uses a residual connection. Mathematically, the residual connection for wider networks can be summarized as: 

\begin{equation}
y = F_{1}(x, \{W_{1_{i}}\}) + F_{2}(x, \{W_{2_{i}}\})
\end{equation}

%The ResNet \cite{szegedy2017inception} architecture is based on residual connections which allows deeper neural networks to be effectively trained. The gradient signal vanishes with increasing network depth. Residual connections are efficient in tackling this problem. The residual block consists of a residual connection which is added after two convolutional blocks. 

%Zagoruyko et al. \cite{zagoruyko2016wide} show that ResNets perform better when they are wider. The paper states that the number of layers need to doubled for an improvement of a fraction of a percent.  Thus, training very deep residual networks has a problem of diminishing feature reuse, which makes these networks very slow to train. To tackle this issue, the paper introduces the idea of using wider networks.

\noindent The functions in equation (1) represent the mapping to be learned. The ``addition" operation refers to an element wise addition with the shortcut connection. The algorithm uses a similar residual connection to connect the second and the fifth layers with the help of a pooling layer i.e., the output of the second layer is added to the output of the 5th layer. The algorithm takes inspiration from the idea in \cite{zagoruyko2016wide} and it introduces a 15 layered convolutional block on the residual connection. With this, the output getting added is the output/feature map of this convolutional block. 

% \subsubsection{Loss Function}
To localize retouching in patches, the model is trained using focal loss, which is mathematically represented in Equation (2). 

\begin{equation}
Focal \space Loss(p_{t}) = -\alpha_{t}(1- p_{t})^{\gamma}\log({p_{t}})
\end{equation}

\begin{equation}
p_{t}= \begin{cases}p & if \ \  y = 0\\1-p & otherwise\end{cases}
\end{equation}

This is derived by modifying the expression for the conventional cross entropy loss. It introduces an additional term of $(1-p_{t})^{\gamma}$ which contains the tunable parameter $\gamma$. The algorithm sets the value of the trainable parameter $\gamma$ to be equal to 5. The benefit of using focal loss is two fold, it helps in localizing objects or regions and addresses large class imbalance. The latter is quite relevant for this application, as retouching or tampering is only present in specific regions. This can lead to large class imbalances in tampered and authentic patches. One of the major advantages of focal loss over cross entropy loss is that focal loss helps in localizing retouched or distinguishing objects/ textures in the patches.  

\subsection{Image Classification}

The classification results on patches are combined to classify the images. The proposed algorithm looks at two methods of classification, namely thresholding and support vector machine (SVM). The predictions are preprocessed to take into account the differences in the number of patches predicted as fake. This is primarily because if retouching is only introduced in specific regions, the difference in the number of authentic patches would not be large. This is even more prevalent in image tampering techniques such as splicing and cloning where only specific regions are tampered. As the CNN network performs better at classifying original patches, we focus on detecting differences in tampered patches. The prediction of the CNN pipeline is therefore post-processed using:

\begin{equation}
\text{Output} = \frac{\text{Total no. of patches predicted as tampered}}{\text{Total  no.  of  patches}}\times100
\end{equation}

The ratio in the equation ensures that for images of different size, a common threshold can be obtained. For large differences in image size, a common threshold of tampered patches would not be able to distinguish whether the difference is due to alterations or size difference. This is the motivation behind using the above mentioned method which normalizes all values for efficient classification. 

For thresholding based classification, the best threshold is searched in the range of 1 to 10 using grid search, and the most optimal threshold is observed to be 4. For decision making, if an image contains more number of tampered patches than the threshold, the image is classified as retouched or fake. Otherwise, it is classified as authentic.   

For SVM based classification, radial basis function kernel is used for training the SVM. The input to the SVM is the output obtained from equation (4). As retouching has been introduced in non-facial regions of the image in the ND-IIITD database \cite{7464282}, all the patches of the retouched images are considered as tampered. For GANs based evaluation, specific regions that are classified using GANs are labeled as tampered. SVM based classification learns the decision boundary based on the training samples and has shown superior results compared to thresholding based approach.

%\newline

\begin{figure}[tbp]

\centerline{\includegraphics[width=0.8\columnwidth, height=5.5cm]{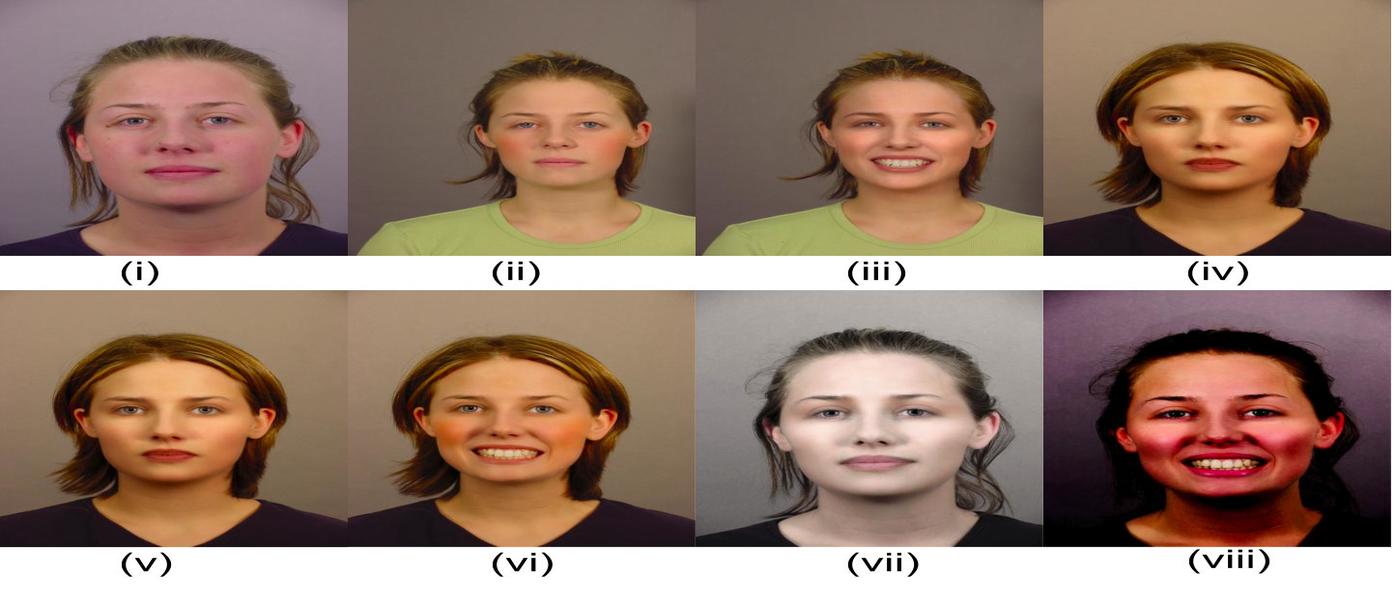}}
\caption{Original and retouched image from each probe. (i) original image; retouched image from (ii) probe 1, (iii) probe 2, (iv) probe 3, (v) probe 4, (vi) probe 5, (vii) probe 6, (viii) probe 7.}
\label{fig2}
\end{figure}

\subsection{Implementation Details}

The model weights are initialized using Xavier's method as shown by Glorot and Bengio \cite{glorot2010understanding}. L1 regularization is used because of its robustness and its ability to select features. Normalization is performed as a preprocessing step to make the data comparable across all the features. To prevent the issue of internal covariate shift while training deep neural networks we normalize the data in each mini-batch. ReLU activation function \cite{xu2015empirical} is used due to its ability to accelerate the convergence of stochastic gradient descent. The CNN network is pre-trained with training data, while testing the test patches are passed through the pre-trained network to get the predictions of the patches. A portion of the training data is used to train the support vector machine and to grid search the threshold value.

\section{Dataset}
\begin{figure*}[t]
  \centering
	%\being{center}
    \hfill
  %\vspace*{20mm}
  \captionbox{Images generated using StarGAN for 9 different attributes. In the figure, H,G, and A refer to hair, gender and age respectively}
  [\linewidth][c]{%
    \includegraphics[width=15cm, height=4cm]{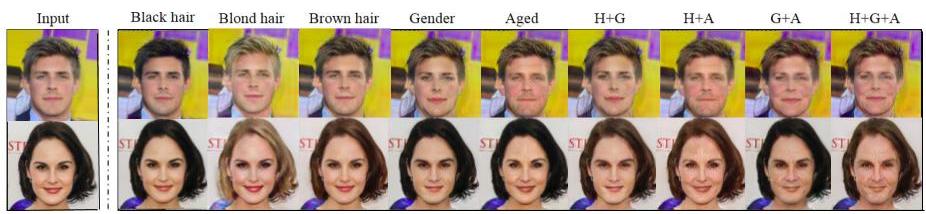}}\quad
  
 \hspace*{\fill}
  %\end{center}
\end{figure*}

%\begin{figure}[t]
%\centering
%\hfill
%\centerline{\includegraphics[width=12.5cm, height=12.5cm]{gans_out3.jpg}}\quad
%\caption{Images generated using StarGAN for 9 different attributes. In the figure, H,G, and A refer to hair, gender and age respectively}
%\label{fig}
%\hspace*{\fill}
%\end{figure}

The paper presents results on two sets of fake images: retouched images and generated images using StarGAN \cite{choi2017stargan}. For the former, the algorithm is trained on the ND-IIITD \cite{7464282} dataset.

\subsection{ND-IIITD Dataset}

The dataset consists of total 4875 face images, out of which 2600 are original collected from the Notre Dame database, Collection B \cite{CollectionB} and 2275 are retouched. The alterations have been introduced using a sophisticated software, PortraitPro Studio Max. There are seven sets of probe images each varying in terms of the characteristics and extent of retouching. In the first two probe sets the level of alterations are lesser compared to other probes and they are only present in local regions. This increases with different probes, with the 7th probe having maximum deviation from the original images. Each probe set contains 325 facial images out of which 211 are males and 114 are females. The following protocols are used for classification:

\begin{enumerate}
\item To be consistent with the protocols followed in literature, 50\% train-test split protocol is followed. The model is tested on 106 males and 57 females from each retouched and original probe. 

\item To learn intra-probe (preset) variations the algorithm is trained on 50\% of the images within a probe (preset) consisting of about 185,000 blocks of size (64,64,3). This is tested on the remaining 163 images of the same preset.

\item To determine whether the trained models are generalizable across different kinds of probe, the model is learned on the 7th probe and tested on the rest. In other words, 325 retouched and 325 original images from probe 7 are used for training, and the remaining dataset is used for testing.   

\end{enumerate}

\subsection{Generated Images}

StarGANs \cite{choi2017stargan} was trained using the CelebA dataset to learn attribute transfer, such as black hair; blond hair; brown hair; gender; aging; hair and gender together; hair and age together; age and gender together; hair, age and gender together. Using StarGANs, 18,000 images are created corresponding to a set of 2000 face images and 9 attributes. Additionally, 15500 original non-overlapping images from the CelebA dataset are used. All the images are of size (128,128,3) and are cropped from the center of the original CelebA dataset images. The GANs model is trained for 20 epochs. Barring some generated images, the rest can deceive the human eye in terms of whether they are generated or original. They are exceedingly similar to original images in terms of facial features, textures and colors. The total database thus comprises 35,500 images. For automated classification of generated images, 2500 images are used for testing, in which 1500 images are from the authentic class and 1000 are from the generated class. Further, 500 images from each class are used for validation. The model is trained on the remaining 32000 images. 

\section{Results}
The result section is divided into three parts according to the experiments. The first subsection discusses the results of detecting facial retouching on the ND-IIITD database and comparison to state-of-the-art reported in literature. The second subsection reports the results of the proposed algorithm in detecting synthetic alterations made using GANs. The third and last subsection examines the impact of image compression on the performance of the proposed algorithm.

\begin{table*}[t]
  \centering
  \caption{Retouching detection accuracy when training and testing sets pertain to the same database \newline }
  \begin{tabular}{|l|l|c|c|c|c|c|c|c|}
  \hline
	
  	\textbf{Input} & \textbf{Method} & \textbf{Probe 1}& \textbf{Probe 2} &\textbf{Probe 3} & \textbf{Probe 4} & \textbf{Probe 5} & \textbf{Probe 6} & \textbf{Probe 7}\\ 
	\hline\hline
	Images & Thresholding & 98.75&98.44&98.44 &96.89&95.11&99.68&96.89 \\
	\hline
	Images & SVM & 97.90 &97.95 &98.96& 97.95&98.75&98.96&93.88  \\ 
	\hline\hline
	Patches & CNN & 99.51 &99.22& 99.02& 99.54 & 99.53 & 99.79 & 99.41  \\ 
	\hline
  \end{tabular}
  \label{tab:1} 
\end{table*}

\subsection{Detecting Facial Retouching}

The proposed architecture with SVM classification yields an overall accuracy of 99.65\% for protocol 1. The class-based accuracies range from 99.38\% to 100\% for probe 1 to 7, respectively. Setting a threshold manually gives slightly lower classification result of 99.48\%. On the same database, Bharati et al.\cite{7464282} achieved 87.1\% and Kee and Farid \cite{Kee19907} achieved 48.8\% accuracy. The same experiment is also performed using image patches of size (64,64,3). It yields an accuracy of 99.42\% with SVM and 99.70\% while using thresholding. These results are summarized in Table 3. In this experiment, all the authentic patches are correctly classified by the CNN architecture. Some of the misclassified patches of size (128,128,3) are shown in Figure 6. For retouched images, the architecture does not perform well on patches which contained either clothing articles or hair. This is because the retouching texture differences gets hidden in clothing or hair texture. %Figure 7 depicts the normalized confusion matrix for patch classification. 

\begin{figure}[tbp]
\centerline{\includegraphics[width=0.8\columnwidth]{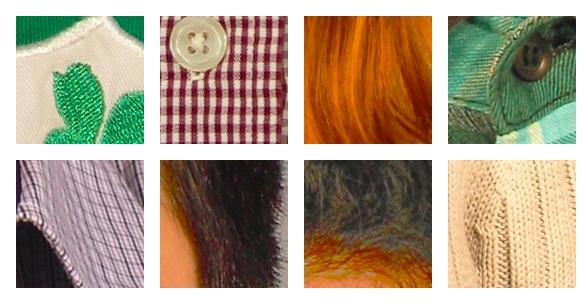}}
\caption{Some examples of misclassified patches}
\label{fig3}
\end{figure}

% \begin{figure}[htbp]

% \centerline{\includegraphics[width=\columnwidth]{Retouching_conf.png}}
% \caption{Normalized confusion matrix for patch predictions using CNN for detecting retouching}
% \label{fig}
% \end{figure}

\begin{table}[]
\begin{center}
\captionof{table}{Normalized confusion matrix for patch predictions using CNN for detecting retouching and generated images.}
\begin{tabular}{ |l|c|c|c|c|} 
\hline
&  \multicolumn{4}{c|}{\textbf{Predicted Labels}} \\
\hline
\multirow{4}{*}{\begin{turn}{-90}\textbf{True Labels} \end{turn}} &  & & Real & Fake  \\ 
\cline{2-5}
& \multirow{2}{*}{\textbf{Retouching}} & Real & 0.99 & 0.01  \\ 
\cline{3-5}
& & Fake & 0.01 & 0.99 \\ 
\cline{2-5}
& \multirow{2}{*}{\textbf{Generated}} & Real & 1.00 & 0.00  \\ 
\cline{3-5}
& & Fake & 0.04 & 0.96 \\ 
\hline
\end{tabular}
\end{center}
\end{table}

% \begin{figure}[htbp]
% \centerline{\includegraphics[width=\columnwidth]{temp.png}}
% \caption{ROC Curve for image classification}
% \label{fig}
% \end{figure}

%To also show that the network wasn’t learning the compression differences of JPG format images present in the dataset, the images were converted to PNG format. The model was fined tuned for the PNG format images. Similar results confirmed that the network wasn’t learning compression. 

The network is tested for its performance in detecting inter-probe and intra-probe variations. For intra-probe variations, protocol 2 is followed and it achieved an accuracy of around 98\% for probes 1,2 and 3; 96\% for probes 4 and 5; and slightly higher accuracy for the probe 6, i.e., ~99\%. However, the predicted accuracy for probe 7 is slightly lower than others. It yields an accuracy of about 96.89\% using thresholding and 93.88\% using SVM for image classification from patches. This shows that the data required to train the model is not very high because of the patch based approach and moreover this performs well with individual probes. The results are summarized in Table 1. The accuracies are lesser as compared to the overall accuracy on the dataset, due to scarcity of data for training for individual probes.  

% \noindent
% \begin{table}[]
% \begin{center}
% \renewcommand\arraystretch{1.5}
% \setlength\tabcolsep{0pt}
% \begin{tabular}{c >{\bfseries}r @{\hspace{0.5em}}c @{\hspace{0.3em}}c @{\hspace{0.5em}}l}
% \captionof{table}{Normalized confusion matrix for patch predictions using CNN for detecting retouching}
%   \multirow{10}{*}{\parbox{0.7cm}{\bfseries\raggedleft actual\\ value}} & 
%     & \multicolumn{2}{c}{\bfseries Prediction outcome} & \\
%   & & \bfseries Real & \bfseries Fake  \\
%   & real$'$ & \MyBox{0.99}{} & \MyBox{0.01}{} \\[2.4em]
%   & fake$'$ & \MyBox{0.01}{} & \MyBox{0.99}{}  \\
% \end{tabular}
% \end{center}
% \end{table}

To test the network for inter-probe variations, one probe set is used for training and the others are used for testing. As probe 7 contains maximum retouching, it is used for training. The overall testing accuracy on probe 1-6 is 99.73\% using thresholding and 99.91\% using SVM.

To assert the importance of the residual connection, it is ablated (i.e. removed). A significant dip in the performance of the model is observed. The patch based accuracy is only 97.83\% and the classification accuracy of the image using thresholding is 96.80\% and 99\% using SVM. The SVM model is finding a high threshold value for distinguishing the two classes as the number of misclassified patches increases. Thus, to ensure a lower false positive rate (FPR), the residual connection is required in the model. 

% \begin{table}[]
% \begin{center}
% \captionof{table}{Accuracies using different patch sizes on the ND-IIITD database for patch classification.}
% \begin{tabular}{ |l|c|c|} 
% \hline
% \textbf{Patch Size} & \textbf{SVM} & \textbf{Thresholding}  \\ 
% \hline
% (64,64,3) & 99.42\% & 99.70\%  \\ 
% \hline
% (128,128,3) & 99.65\% & 99.48\% \\ 
% \hline
% %\multicolumn{3}{l}{TABLE II: Accuracies using different patch size on ND-IIITD database}
% \end{tabular}
% \end{center}
% \end{table}

\begin{table}[]
\begin{center}
\caption{Overall image retouching detection accuracy and comparison with existing reported results in literature}
\begin{tabular}{ |l|c| } 
\hline
\textbf{Algorithm} & \textbf{Accuracy}\\\hline
Kee and Farid \cite{Kee19907} & 48.8\%  \\ 
\hline
Bharati (Unsupervised DBM) \cite{7464282} & 81.9\%  \\ 
\hline
Bharati (Supervised DBM) \cite{7464282} & 87.1\%  \\ 
\hline
Proposed (Thresholding) - (64,64,3) & 99.70\%  \\ 
\hline
Proposed (SVM) - (64,64,3) & 99.42\%  \\ 
\hline
Proposed (Thresholding) - (128,128,3) & 99.48\%  \\ 
\hline
Proposed (SVM) - (128,128,3) & 99.65\%  \\ 
\hline
\end{tabular}
\end{center}
\end{table}

%\vspace*{-10mm}

\subsection{Detecting Generated Images}

The overall accuracy of the proposed algorithm on generated images from STARGAN \cite{choi2017stargan} is 99.83\% using thresholding and 99.73\% using support vector machine for image classification from patches. The images are converted to JPG format to test the models performance in presence of lossy compression. The images are compressed using a quality factor of 50. The patch classification accuracy on compressed images is 95.6\% and image classification using SVM yields an accuracy of 96.33\%. The accuracy for JPG compressed images using thresholding is significantly lesser, 88.89\%. One of the major reason is the higher false positive rate. This also shows the added advantage of using support vector machines for predicting the labels of images as generated or real. These results are summarized in Table 4 below. Figure 7 shows some examples for patches which are correctly and falsely classified. Table 5 compares the performance of proposed algorithm for detecting synthetic alterations with Bharati et al. \cite{7464282} which yields an accuracy of 91.83\%.  
%There are 3 authentic patches which are misclassified and are shown in figure 8. Figure 9 shows the correlation matrix plot for patch based classification from the convolutional neural network. 
%As the image size is (128,128,3), therefore, there are only 4 patches present in each image. This is another reason for the poor performance of the model using thresholding for JPG compressed images.

\begin{figure}[tbp]
\centerline{\includegraphics[width=0.8\columnwidth]{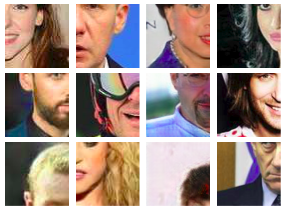}}
\caption{Column 1: Correctly classified authentic patches; Column 2: Correctly classified generated patches; Column 3: Misclassified authentic patches; Column 4: Misclassified generated patches}
\label{fig4}
\end{figure}

% \begin{figure}[tbp]
% \centerline{\includegraphics[width=1.2\columnwidth]{GANS_conf.png}}
% \vspace{6pt}
% \caption{Normalized confusion matrix for patch predictions using CNN for detecting generated images}
% \label{fig}
% \end{figure}

% % \noindent
% \renewcommand\arraystretch{1.5}
% \setlength\tabcolsep{0pt}
% \begin{tabular}{c >{\bfseries}r @{\hspace{0.5em}}c @{\hspace{0.3em}}c @{\hspace{0.5em}}l}
%   \multirow{10}{*}{\parbox{0.7cm}{\bfseries\raggedleft actual\\ value}} & 
%     & \multicolumn{2}{c}{\bfseries Prediction outcome} & \\
%   & & \bfseries Real & \bfseries Fake  \\
%   & real$'$ & \MyBox{1.00}{} & \MyBox{0.00}{} \\[2.4em]
%   & fake$'$ & \MyBox{0.04}{} & \MyBox{0.96}{}  \\
% \end{tabular}

% \begin{table}[]
% \begin{center}
% \captionof{table}{Normalized confusion matrix for patch predictions using CNN for detecting generated images. \newline}
% \begin{tabular}{ |l|c|c|c|} 
% \hline
% &  \multicolumn{3}{c|}{Predicted Labels} \\
% \hline
% %\multirow{4}{*}{\begin{turn}{-90} True Labels \end{turn}} & & Real & Fake  \\ 
% True Labels & & Real & Fake \\
% % & & &  \\
% % &  \multicolumn{3}{|c|}{} \\
% & Real & 1.00 & 0.00 \\ 
% % & & &  \\
% % &  \multicolumn{3}{|c|}{} \\
% & Fake & 0.04 & 0.96 \\ 
% \hline
% %\multicolumn{3}{l}{TABLE II: Accuracies using different patch size on ND-IIITD database}
% \end{tabular}
% \end{center}
% \end{table}

\begin{table}[]
\begin{center}
\caption{Classification accuracies for generated images for different image formats}
\begin{tabular}{ |c|c|c|} 
\hline
\textbf{Compression}  & \textbf{SVM} & \textbf{Thresholding}  \\ 
\hline
PNG images & 99.73\% & 99.83\%  \\ 
\hline
JPG images & 96.33\% & 88.89\% \\ 
\hline
\end{tabular}
\end{center}
\end{table}

\begin{table}[]
\begin{center}
\caption{Overall generated image detection accuracy and comparison with other algorithms for PNG format images}
\begin{tabular}{ |l|c| } 
\hline
\textbf{Algorithm} & \textbf{Overall Accuracy}  \\ 
\hline
Bharati \cite{7464282} & 91.83\%  \\
\hline
Proposed (Thresholding) & 99.83\%  \\ 
\hline
Proposed (SVM) & 99.73\%  \\ 
\hline
\end{tabular}
\end{center}
\end{table}

\subsection{Compression Analysis for Retouching}

Deep learning models perform well in detecting double compressed JPG images \cite{jpg_compression, jpg}. Introducing tampering/retouching in an already compressed JPG image followed by re-compression afterwards leads to double compression. To ensure that the model is not learning such variations, JPG format images present in the retouching dataset are converted to PNG format and the model is fine tuned for PNG format images. Similar results confirmed that the network is not learning the properties of image compression.

\section{Conclusion and Future Work}
The paper presents a convolutional neural network architecture for detecting digital manipulations in terms of retouching and GANs based alterations. The results are demonstrated on two databases: ND-IIITD database and images generated using StarGANs. The proposed algorithm shows significant improvements compared to the results reported in the literature. This paper also opens another possible avenue for research in classification of generated images and testing photorealism as well. While this paper analyzes the images generated using StarGANs, the idea of unified automatic detection could be an interesting extension. %\cite{Authors11}

\section{Acknowledgement}
Vatsa and Singh are partly supported through Infosys Center for Artificial Intelligence at IIIT-Delhi. The research is also partly supported through a grant from MEITY, Government of India.

{\small
\bibliographystyle{ieee}
\bibliography{submission_example}

\begin{thebibliography}{10}\itemsep=-1pt

\bibitem{image2}
{Age Defying Techniques}.
\newblock \url{https://bit.ly/2LzDpLa}.
\newblock Accessed: 2018-04-26.

\bibitem{jennifer}
{Jennifer Lopez Retouched Image}.
\newblock \url{https://bit.ly/2NWdn1A}.
\newblock Accessed: 2018-04-26.

\bibitem{photoshop}
{New Israeli law bans use of too-skinny models in ads}.
\newblock \url{https://cnn.it/1mNTiY1}.
\newblock Accessed: 2018-04-26.

\bibitem{ralph}
{Ralph Lauren} apologises for digitally retouching slender model to make her
  head look bigger than her waist.
\newblock \url{https://dailym.ai/1zOvgAh}.
\newblock Accessed: 2018-04-18.

\bibitem{7464282}
A.~Bharati, R.~Singh, M.~Vatsa, and K.~W. Bowyer.
\newblock Detecting facial retouching using supervised deep learning.
\newblock {\em IEEE Transactions on Information Forensics and Security},
  11(9):1903--1913, Sept 2016.

\bibitem{DBLP:journals/corr/abs-1709-07598}
A.~Bharati, M.~Vatsa, R.~Singh, K.~W. Bowyer, and X.~Tong.
\newblock Demography-based facial retouching detection using subclass
  supervised sparse autoencoder.
\newblock {\em CoRR}, abs/1709.07598, 2017.

\bibitem{choi2017stargan}
Y.~Choi, M.~Choi, M.~Kim, J.-W. Ha, S.~Kim, and J.~Choo.
\newblock Stargan: Unified generative adversarial networks for multi-domain
  image-to-image translation.
\newblock {\em arXiv preprint arXiv:1711.09020}, 2017.

\bibitem{impact_biometric}
M.~Ferrara, A.~Franco, D.~Maltoni, and Y.~Sun.
\newblock On the impact of alterations on face photo recognition accuracy.
\newblock In A.~Petrosino, editor, {\em Image Analysis and Processing}, pages
  743--751. Springer Berlin Heidelberg, 2013.

\bibitem{CollectionB}
P.~J. Flynn, K.~W. Bowyer, and P.~J. Phillips.
\newblock Assessment of time dependency in face recognition: An initial study.
\newblock In J.~Kittler and M.~S. Nixon, editors, {\em Audio- and Video-Based
  Biometric Person Authentication}, pages 44--51. Springer Berlin Heidelberg,
  2003.

\bibitem{glorot2010understanding}
X.~Glorot and Y.~Bengio.
\newblock Understanding the difficulty of training deep feedforward neural
  networks.
\newblock In {\em Proceedings of the thirteenth international conference on
  artificial intelligence and statistics}, pages 249--256, 2010.

\bibitem{pix2pix}
P.~Isola, J.~Zhu, T.~Zhou, and A.~A. Efros.
\newblock Image-to-image translation with conditional adversarial networks.
\newblock {\em CoRR}, abs/1611.07004, 2016.

\bibitem{Kee19907}
E.~Kee and H.~Farid.
\newblock A perceptual metric for photo retouching.
\newblock {\em Proceedings of the National Academy of Sciences},
  108(50):19907--19912, 2011.

\bibitem{kose}
N.~Kose, L.~Apvrille, and J.~L. Dugelay.
\newblock Facial makeup detection technique based on texture and shape
  analysis.
\newblock In {\em 11th IEEE International Conference and Workshops on Automatic
  Face and Gesture Recognition (FG)}, May 2015.

\bibitem{jpg_compression}
B.~Li, H.~Luo, H.~Zhang, S.~Tan, and Z.~Ji.
\newblock A multi-branch convolutional neural network for detecting double
  {JPEG} compression.
\newblock {\em CoRR}, abs/1710.05477, 2017.

\bibitem{russello2009impact}
S.~Russello.
\newblock The impact of media exposure on self-esteem and body satisfaction in
  men and women.
\newblock {\em Journal of Interdisciplinary Undergraduate Research}, 1(1):4,
  2009.

\bibitem{singh2013face}
A.~Singh, S.~Tiwari, and S.~K. Singh.
\newblock Face tampering detection from single face image using gradient
  method.
\newblock {\em International Journal of Security and its Applications}, 7(1),
  2013.

\bibitem{szegedy2017inception}
C.~Szegedy, S.~Ioffe, V.~Vanhoucke, and A.~A. Alemi.
\newblock Inception-v4, inception-resnet and the impact of residual connections
  on learning.
\newblock In {\em AAAI}, volume~4, page~12, 2017.

\bibitem{jpg}
Q.~Wang and R.~Zhang.
\newblock Double jpeg compression forensics based on a convolutional neural
  network.
\newblock {\em EURASIP Journal on Information Security}, 2016(1):23, Oct 2016.

\bibitem{xu2015empirical}
B.~Xu, N.~Wang, T.~Chen, and M.~Li.
\newblock Empirical evaluation of rectified activations in convolutional
  network.
\newblock {\em arXiv preprint arXiv:1505.00853}, 2015.

\bibitem{zagoruyko2016wide}
S.~Zagoruyko and N.~Komodakis.
\newblock Wide residual networks.
\newblock {\em arXiv preprint arXiv:1605.07146}, 2016.

\bibitem{cyclegans}
J.~Zhu, T.~Park, P.~Isola, and A.~A. Efros.
\newblock Unpaired image-to-image translation using cycle-consistent
  adversarial networks.
\newblock {\em CoRR}, abs/1703.10593, 2017.

\end{thebibliography}
}

\end{document}